\documentclass[letterpaper, 10 pt, conference]{ieeeconf}  

\IEEEoverridecommandlockouts                              

\usepackage{graphics} 
\usepackage{epsfig} 
\usepackage{mathptmx} 
\usepackage{times} 
\usepackage{amsmath} 
\usepackage{amssymb}  

\usepackage{graphicx} 
\usepackage{amsmath} 
\usepackage{caption} 
\usepackage{algorithm} 
\usepackage{algorithmic} 

\usepackage{booktabs} 
\usepackage{siunitx} 
\usepackage{bm} 

\title{\LARGE \bf
FlightDiffusion: Revolutionising Autonomous Drone Training with Diffusion Models Generating FPV Video
}

\author{Valerii Serpiva$^{*}$, Artem Lykov$^{*}$, Faryal Batool, Vladislav Kozlovskiy,\\Miguel Altamirano Cabrera and Dzmitry Tsetserukou 
\thanks{$^{*}$ These authors contributed equally to this work.}
\thanks{The authors are with the Intelligent Space Robotics Laboratory, Skolkovo Institute of Science and Technology Moscow, Bolshoy Boulevard 30, bld. 1, 121205, Moscow, Russia.
\tt \{Valerii.Serpiva, Artem.Lykov, Faryal.Batool, Vladislav.Kozlovskiy, M.Altamirano, D.Tsetserukou\}@skoltech.ru}
}

\begin{document}

\maketitle
\thispagestyle{empty}
\pagestyle{empty}

\begin{abstract}

We present FlightDiffusion, a diffusion-model-based framework for training autonomous drones from first-person view (FPV) video. Our model generates realistic video sequences from a single frame, enriched with corresponding action spaces to enable reasoning-driven navigation in dynamic environments. Beyond direct policy learning, FlightDiffusion leverages its generative capabilities to synthesize diverse FPV trajectories and state-action pairs, facilitating the creation of large-scale training datasets without the high cost of real-world data collection. Our evaluation demonstrates that the generated trajectories are physically plausible and executable, with a mean position error of 0.25 m (RMSE 0.28 m) and a mean orientation error of 0.19 rad (RMSE 0.24 rad). This approach enables improved policy learning and dataset scalability, leading to superior performance in downstream navigation tasks. Results in simulated environments highlight enhanced robustness, smoother trajectory planning, and adaptability to unseen conditions. An ANOVA revealed no statistically significant difference between performance in simulation and reality (F(1, 16) = 0.394, p = 0.541), with success rates of M = 0.628 (SD = 0.162) and M = 0.617 (SD = 0.177), respectively, indicating strong sim-to-real transfer. The generated datasets provide a valuable resource for future UAV research. This work introduces diffusion-based reasoning as a promising paradigm for unifying navigation, action generation, and data synthesis in aerial robotics.

\end{abstract}

\section{Introduction}


The field of autonomous navigation for Unmanned Aerial Vehicles (UAVs) has revolutionized applications ranging from search and rescue \cite{drones7030202}, to autonomous drone racing \cite{Scaramuzza}, and rapid response in unstructured environments \cite{drones5030057}. A key barrier to deploying these systems robustly in complex, dynamic settings \cite{GUO2021479} is the challenge of developing navigation policies that can reason about unseen situations and adapt in real-time. Traditional data-driven methods, such as Reinforcement Learning (RL) \cite{hodge2021deep} and Imitation Learning (IL) \cite{wang2021dynamic}, have shown promise but are notoriously sample-inefficient. They require vast and diverse datasets of flight experience, which are logistically complex, expensive, and often dangerous to collect in the real world. This reliance on large-scale data creates a significant bottleneck for innovation and deployment.

To address these challenges, the field is increasingly turning towards generative models. Recent works have demonstrated the power of diffusion models to learn complex quadrotor dynamics \cite{das2025dronediffusionrobustquadrotordynamics} and generate specialized aerobatic trajectories \cite{zhong2025automaticgenerationaerobaticflight}. Concurrently, the integration of Large Language Models (LLMs) and Vision-Language Models (VLMs) is endowing drones with high-level semantic reasoning. These models enable UAVs to understand natural language commands \cite{Liu_2023_ICCV}, perform sophisticated path planning \cite{xiao2025fmplannerfoundationmodelguided}, and dynamically adapt their behavior based on contextual understanding \cite{Khan2025_ContextAwareDrone, 10967934}. This has led to more intelligent and human-responsive systems for specialized tasks like drone racing and large-scale mission execution \cite{10974117}.

\begin{figure}[t]
\centering
\includegraphics[width=1.0\linewidth]{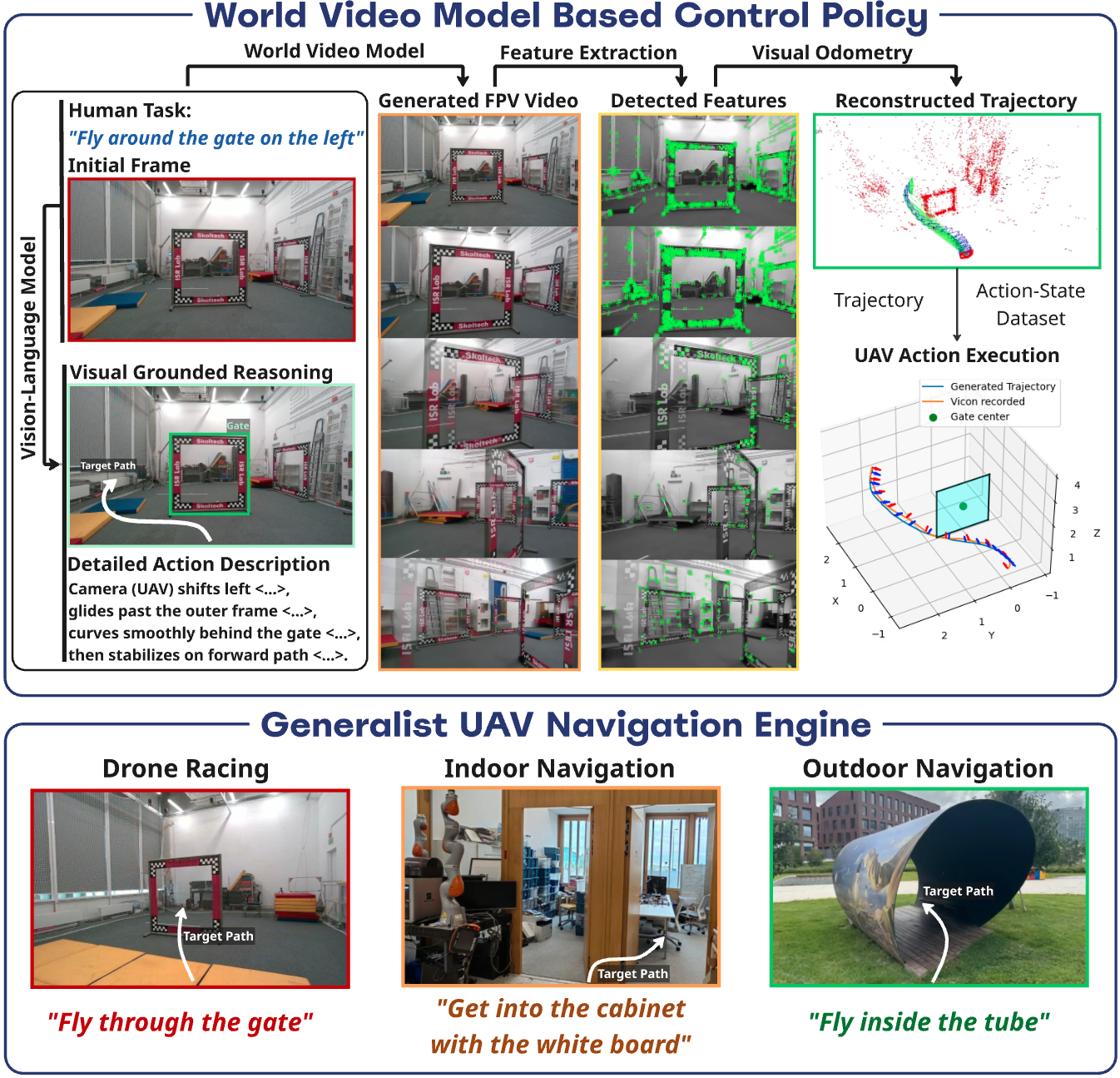}
\caption{FlightDiffusion System Overview. The framework integrates a reasoning module and a generalization module to process a single input frame. A video diffusion model then generates future frames and their corresponding action spaces. A visual odometry module ensures the physical plausibility of the synthesized trajectories.}
\label{fig:teaser}
\end{figure}

However, a gap remains between high-level semantic planning and low-level, physically-grounded policy execution. While VLMs can determine what to do, generating the precise, continuous visuomotor controls for how to do it remains a data-intensive problem. This paper introduces FlightDiffusion, a novel framework that bridges this gap by unifying high-level reasoning, generative synthesis of first-person view (FPV) video, and action-rich dataset creation. From a single starting frame, FlightDiffusion generates realistic, physically plausible video sequences of flight trajectories. Crucially, it enriches these videos with their corresponding action spaces, creating a powerful tool for reasoning-driven navigation.

Our contributions are as follows:
\begin{enumerate}
\item We propose a diffusion-model-based framework that generates realistic, long-horizon FPV flight videos conditioned on a high-level plan.
\item We demonstrate that our model can simultaneously generate a corresponding action space.
\item We leverage FlightDiffusion as a data engine to synthesize diverse, action-rich FPV datasets, mitigating the high cost of real-world data collection and enabling more robust training of downstream navigation policies.
\end{enumerate}

Our approach underwent a comprehensive validation framework, integrating both simulated and physical real-world testing environments. The evaluation metrics focused on two primary aspects: the precision of visual odometry tracking applied to synthetically generated videos, and the overall efficacy of autonomous task execution. Empirical results confirm that control policies trained using our synthesized visual trajectories exhibit significantly enhanced robustness, improved smoothness in planned trajectories, and stronger generalization capabilities when deployed under previously unseen environmental conditions.

\section{Related Works}

Autonomous navigation has long been a central challenge in robotics, as it underpins applications ranging from inspection and agriculture to racing and disaster response. A vast body of research has explored different approaches to this problem, including control-theoretic methods, data-driven learning, and more recently generative and foundation models. Despite these advances, defining a navigation policy that can generalize across tasks, environments, and embodiments remains an open and critical goal.

\subsection{Data-Driven UAV Navigation}

Autonomous navigation for UAVs has long relied on data-driven methods such as Reinforcement Learning (RL) and Imitation Learning (IL). RL enables policies to emerge from trial-and-error exploration \cite{sutton2018reinforcement}, but it remains sample-inefficient and faces persistent sim-to-real transfer issues. IL offers a more direct route by leveraging expert demonstrations \cite{osa2018algorithmic}, yet its scalability is constrained by the expense of collecting large amounts of high-quality data. Recent advances attempt to overcome these barriers. Tayar et al. \cite{tayar2025autonomousuavflightnavigation} demonstrated RL-based UAV navigation in cluttered, GPS-denied spaces, showing adaptability but at the cost of intensive data requirements. Similarly, Wei et al. \cite{Wei_2025} proposed an imitation learning framework for orchard environments that uses a VAE-based controller to map raw RGB inputs directly to controls, reducing reliance on explicit mapping but still requiring repeated expert interventions. These works underscore the need for scalable alternatives to raw experience collection. FlightDiffusion addresses this bottleneck by generating unlimited, action-annotated FPV flight data, sidestepping the logistical and safety challenges of real-world data collection.

\subsection{Generative Models for UAV Dynamics}

Generative models, particularly diffusion models, have recently emerged as a promising paradigm in aerial robotics. DroneDiffusion has been used to learn quadrotor dynamics directly from data, offering improvements over classical models \cite{das2025dronediffusionrobustquadrotordynamics}, while Zhong et al. \cite{zhong2025automaticgenerationaerobaticflight} applied diffusion to synthesize aerobatic trajectories beyond the reach of conventional planners. Constraint-guided extensions such as CGD \cite{kondo2024cgdconstraintguideddiffusionpolicies} further integrate safety and feasibility constraints, ensuring generated UAV trajectories remain dynamically valid and collision-free. In parallel, Zhao et al. \cite{zhao2025improvingdroneracingperformance} showed how iterative learning MPC can enhance racing performance by refining predictions across laps, emphasizing the importance of trajectory adaptation. Collectively, these approaches highlight diffusion’s ability to create safe and diverse flight paths. However, they remain focused on state or trajectory sequences. FlightDiffusion departs from this trajectory-only view by synthesizing FPV video sequences enriched with corresponding action spaces, creating a sensor-grounded data engine that links perception and control more directly.

\subsection{Foundation Models for UAV Intelligence}

The integration of foundation models has advanced UAVs toward semantic and contextual reasoning. Vision-language and large language models have been used to translate high-level instructions into executable goals \cite{Liu_2023_ICCV, xiao2025fmplannerfoundationmodelguided, Khan2025_ContextAwareDrone, 10967934}. Zhang et al. \cite{zhang2025embodiednavigationfoundationmodel} introduced NavFoM, a unified foundation model trained across multiple robot embodiments and tasks, demonstrating robust generalization without task-specific tuning. Ye et al. \cite{VLM-RRT} proposed VLM-RRT, which guides sampling-based motion planning with semantic cues from language, showing how VLMs can inject task awareness into low-level planners. Similarly, Suomela et al. \cite{suomela2025syntheticvsrealtraining} compared synthetic and real training data for navigation, highlighting the persistent sim-to-real gap and the value of hybrid strategies. These works illustrate how foundation models can reason about \textit{what} to do, but they typically rely on hand-crafted or existing controllers for \textit{how} to do it. By contrast, FlightDiffusion provides a generative bridge: conditioning on high-level prompts, it produces realistic FPV video and corresponding action sequences that ground semantic plans in visuomotor experience, enabling policies that generalize with fewer real-world samples.


\section{System Architecture}

The FlightDiffusion system employs a modular, sequential pipeline comprising three core stages: Visual Reasoning and Mission Planning, Video Generation and 3D Trajectory Reconstruction (Fig. \ref{fig:teaser}). 

The first module is responsible for interpreting the scene and generating high-level, semantically-rich commands. We leverage Gemini 2.5 Flash \cite{deepmind2024gemini}, a powerful multimodal Vision-Language Model (VLM) from Google DeepMind, for this task. The pipeline is initiated by a single RGB image, denoted as $I_{input}$, which is captured from an onboard drone camera. This image is then standardized into a tensor $I_{std}$ passed to is fed into the Gemini API alongside a structured natural language prompt engineered to elicit specific reasoning about the environment and desired drone maneuver. This approach allows Gemini 2.5 Flash to perform long-horizon planning tasks implicitly by understanding complex scene geometry and objects (e.g., gates, obstacles, windows). The output of this module is a text string, $C_{mission}$ which encapsulates the high-level mission plan.

The core task of visual trajectory generation is achieved through a powerful generative model. This stage takes the standardized image $I_{std}$ and the mission command $C_{mission}$ as conditional inputs. Our approach leverages the conditioning mechanism inherent in contemporary image-to-video (I2V) diffusion models, which generate video sequences from an initial frame and a text prompt. This capability is now standard in state-of-the-art systems such as Luma Ray2 \cite{luma_ray2_2025}, Google Veo 2 \cite{veo2_google2024}, Seedance 1.0 Pro \cite{seedance2024}, and Wan 2.2 \cite{wan22_replicate2025}. For the purposes of this study, we selected the Wan 2.2 I2V Fast model, prioritizing its operational efficiency and satisfactory output fidelity. This model was chosen for its ability to generate high-quality, temporally consistent short video $V_{\text{gen}}$ conditioned on a starting image and a text prompt describing the desired motion (e.g., "a drone flying forward through a gate"). The average video generation time varies from 20 to 60 sec.

\begin{figure}[t]
\centering
\includegraphics[width=1.0\linewidth]{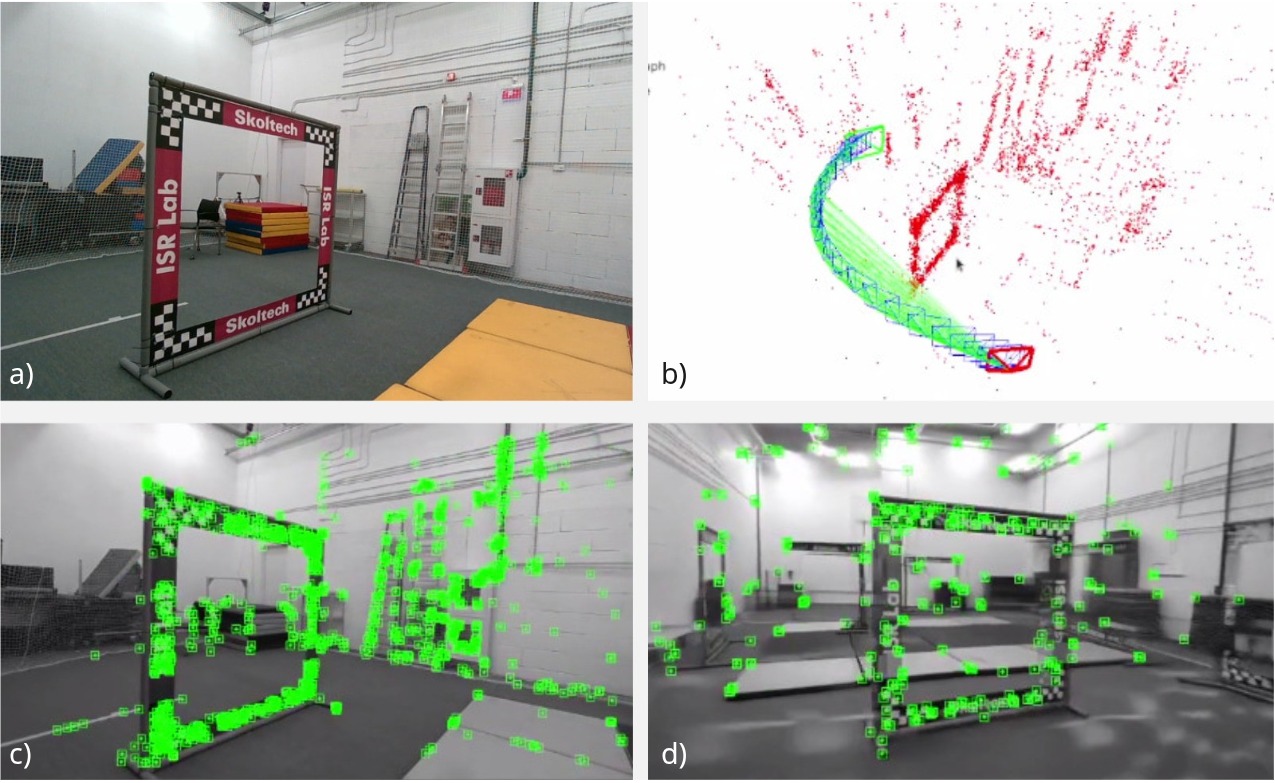}
\caption{Visualization of the visual odometry pipeline for synthetic video generation, showing the sequential frame processing used for camera pose estimation and trajectory reconstruction: a) initial frame, b) ORB-SLAM tracking, c) features extracted from the first frame, and d) features in the last generated frame.}
\label{fig:orb_slam}
\end{figure}

The generated video $V_{\text{gen}}$ is subsequently processed by a visual odometry module in Monocular mode, specifically ORB-SLAM3~\cite{ORBSLAM3_TRO}, to extract a precise representation of the action-state space (Fig. \ref{fig:orb_slam}). This data, which includes the drone's estimated 3D trajectory, velocity, and pose relative to the environment, are then used to train the control policies of an autonomous drone. In the final operational stage, the designed drone performs these actions in the real world, with its onboard camera providing a live visual feed, effectively closing the loop between synthetic training and real world execution.

\section{Skill Repetition from Generated Videos}

The Reasoning Block serves as the cognitive core of the FlightDiffusion system. Its primary function is to act as a high-level mission planner, translating a concise user command and a single environmental image into a detailed, executable flight plan. The block operates by leveraging a multi-modal LLM. The process is as follows:

\begin{itemize}
\item \textbf{Input Processing}: The block accepts two key inputs: a static image from the drone's onboard camera (or a simulated starting point) and a natural language task description (e.g.,``enter the restaurant through the door").

\item \textbf{Contextual Analysis}: The image is encoded and sent to the LLM alongside two meticulously crafted text prompts: a system prompt that defines the agent's role as a drone navigation expert and specifies the required output format, and a user prompt that incorporates the task description.

\item \textbf{Mission Generation}: The multi-modal LLM analyzes the visual scene (identifying objects, obstacles, and landmarks) and synthesizes it with the task goal. It then generates a comprehensive, step-by-step textual description of the optimal flight path. This description includes details such as approach vectors, waypoints, relative velocities, and specific maneuvers, effectively creating a ``recipe'' for the flight.

\end{itemize}

The concept of Skill Repetition from Generated Videos is achieved by using the output of this pipeline as a synthetic training environment for the drone's control systems. The process for skill acquisition is:

\begin{itemize}
\item \textbf{Video and Trajectory Generation}: The detailed mission description from the Reasoning Block is used to condition a video diffusion model, which generates a photorealistic FPV video simulating the successful execution of the planned flight.

\item \textbf{State-Action Extraction}: This generated video is then processed by a visual odometry framework (e.g., ORB-SLAM3). This framework analyzes the sequence of video frames to reconstruct the camera's (and thus the drone's) precise 3D trajectory, estimating its pose, velocity, and acceleration at every timestep.

\item \textbf{Policy Training}: The extracted data forms a rich, supervised training dataset of state-action pairs. Each state is the visual observation and estimated pose, and each action is the corresponding movement that led to the next state. A drone's control policy (e.g., a neural network controller) can be trained via imitation learning or reinforcement learning to replicate these actions from the corresponding states.

\item \textbf{Repetition and Mastery}: By generating thousands of variations of a specific skill (e.g., ``door entry") from different starting images and perspectives, the drone can be exposed to a vast and diverse set of scenarios. Its policy repeatedly practices the skill in simulation, learning robust and generalizable maneuvers without the risk, cost, and time associated with real-world flight trials.

\end{itemize}


\section{Trajectory, Action-Space Extraction and Mission Execution}

Although ORB-SLAM3 can deliver accurate and reliable trajectories, its full mapping and relocalization features are often unnecessary and computationally heavy when the task is short-horizon, goal-driven navigation guided by a pre-recorded video. In our approach, the main idea is to take a first-person video sequence generated, $V_{\text{gen}} = {I_0, I_1, ..., I_N}$, where $I_t$ denotes the synthetic frame at time $t$, and convert it into a practical short 3D trajectory.

The process can be formalized as a function $F$ mapping the video to a trajectory:
\begin{equation}
T_{\text{world}}^{\text{cam}}(t) = F(V_{\text{gen}}; \Theta_{\text{SLAM}})
\end{equation}
where $T_{\text{world}}^{\text{cam}}(t) \in SE(3)$, is the estimated camera pose (a rigid body transformation matrix) for frame $I_t$, and $\Theta_{\text{SLAM}}$ represents the parameters and state of the SLAM system (e.g., the map, keyframes, covisibility graph).

For each new synthetic frame $I_t$, ORB-SLAM3 performs the following key steps:

\begin{enumerate}
    \item \textbf{Feature Extraction and Matching:} ORB features are extracted from $I_t$, resulting in a set of keypoints $K_t = \{\mathbf{k}_i\}$ with descriptors. These are matched against features in the current map $M$ or a reference keyframe.
    \item \textbf{Pose Optimization:} The camera pose is estimated by minimizing the reprojection error between matched 3D map points $\mathbf{X}_j \in \mathbb{R}^3$, and their corresponding 2D keypoints $\mathbf{k}_j$.
    \begin{equation}
    T^* = \arg\min_{T \in SE(3)} \sum_j \rho \left( \left\| \pi(T, \mathbf{X}_j) - \mathbf{k}_j \right\|^2_2 \right)
    \label{eq:pose_optim}
    \end{equation}
    where $\pi(\cdot)$ is the camera projection function and $\rho$ is a robust kernel (e.g., Huber) to mitigate the effect of outliers.
    \item \textbf{Map Point Triangulation and Bundle Adjustment:} New map points are triangulated from matched features between keyframes. Local Bundle Adjustment optimizes both the camera poses of a set of keyframes and the positions of the observed map points to achieve a globally consistent sparse reconstruction.
\end{enumerate}

The output is a sequence of optimized camera poses $\{T_0, T_1, ..., T_N\}$ which form the trajectory $T_{\text{world}}^{\text{cam}}(t)$ and $\text{State-Actions pairs}(\bm{\xi}, \mathbf{v})$ that the drone is commanded to follow.

\begin{algorithm}[H]
\caption{Trajectory Extraction from Synthetic Video}
\begin{algorithmic}[1]
\STATE \textbf{Input:} Generated video $V_{\text{gen}}$
\STATE \textbf{Output:} Estimated trajectory $\{T_0, T_1, ..., T_N\}$, $\text{State-Actions}(\bm{\xi}, \mathbf{v})$
\STATE Initialize ORB-SLAM3 system (Monocular mode)
\FOR{each frame $I_t$ in $V_{\text{gen}}$}
    \STATE $T_t \leftarrow \text{ORB-SLAM3.Track}(I_t)$ \hfill $\triangleright$ Solve Eq.~\ref{eq:pose_optim}
    \STATE Add $T_t$ to trajectory
    \STATE $\text{StorePair}(\bm{\xi}_{\text{current}}, \mathbf{v}^c_t)$ \hfill $\triangleright$ Record the state-action pair
\ENDFOR
\STATE \textbf{return} trajectory
\end{algorithmic}
\label{alg:traj_extract}
\end{algorithm}

The physical drone's state is continuously estimated in real-time using the OpenVINS\cite{9196524} pipeline. The state of the drone at time $t$ is defined by its pose in the world frame:
\begin{equation}
\bm{\xi}_t = \left[ \mathbf{p}_t, \mathbf{q}_t \right]^T ,
\end{equation}
where $\mathbf{p}_t = [x, y, z]^T \in \mathbb{R}^3$, is the positional coordinate and $\mathbf{q}_t$ is the unit quaternion representing the orientation. 

The trajectory $\{T_0, T_1, ..., T_N\}$ extracted from the synthetic video consists of a sequence of desired future poses, $T_i^{\text{des}} \in SE(3)$. This trajectory is transformed into a velocity command for the drone's flight controller.

Finally, the process can be formalized as a function $\mathcal{F}$ mapping the video to a trajectory:
\begin{equation}
\{T_0^{\text{des}}, T_1^{\text{des}}, ..., T_N^{\text{des}}\} = {F}(V_{\text{gen}}; \Theta_{\text{SLAM}}) ,
\end{equation}
where $T_i^{\text{des}} \in SE(3)$ is the desired camera pose for the $i$-th step. This trajectory is then passed to the controller described to generate the velocity commands $\mathbf{v}^c$ that guide the physical drone. The full navigation pipeline is given in Algorithm \ref{alg:full_pipeline}.

\begin{algorithm}[H]
\caption{Full Navigation Pipeline}
\begin{algorithmic}[1]
\STATE \textbf{Input:} Global Task Description, Initial Real Image $I_0^{\text{real}}$
\STATE Reason about task and generate subtask list $L$
\STATE \textbf{for each} subtask $s_j$ \textbf{in} $L$ \textbf{do}
\STATE \quad Reason $s_j$ and image $I^{\text{real}}$
\STATE \quad Generate synthetic video $V_{\text{gen}}$ for $s_i$
\STATE \quad Extract desired trajectory $\{T_i^{\text{des}}\}$ from $V_{\text{gen}}$
\STATE \quad \textbf{for each} $T_i^{\text{des}}$ \textbf{in} trajectory \textbf{do}
\STATE \qquad \textbf{while} drone has not reached $T_i^{\text{des}}$ \textbf{do}
\STATE \qquad \quad Estimate current state $\bm{\xi}_t$
\STATE \qquad \quad Calculate velocity command $\mathbf{v}^c$
\STATE \qquad \quad Send $\mathbf{v}^c$ to flight controller
\STATE \qquad \textbf{end while}
\STATE \quad Take a new Image $I_j^{\text{real}}$
\STATE \quad \textbf{end for}
\STATE \textbf{end for}
\end{algorithmic}
\label{alg:full_pipeline}
\end{algorithm}

\section{Experimental Evaluation}

This section details the experimental setup, methodology, and results of evaluating FlightDiffusion, our framework for autonomous drone navigation using diffusion-based FPV video generation.

\subsection{Trajectory Generation}

\begin{figure}[t]
\centering
\includegraphics[width=1.0\linewidth]{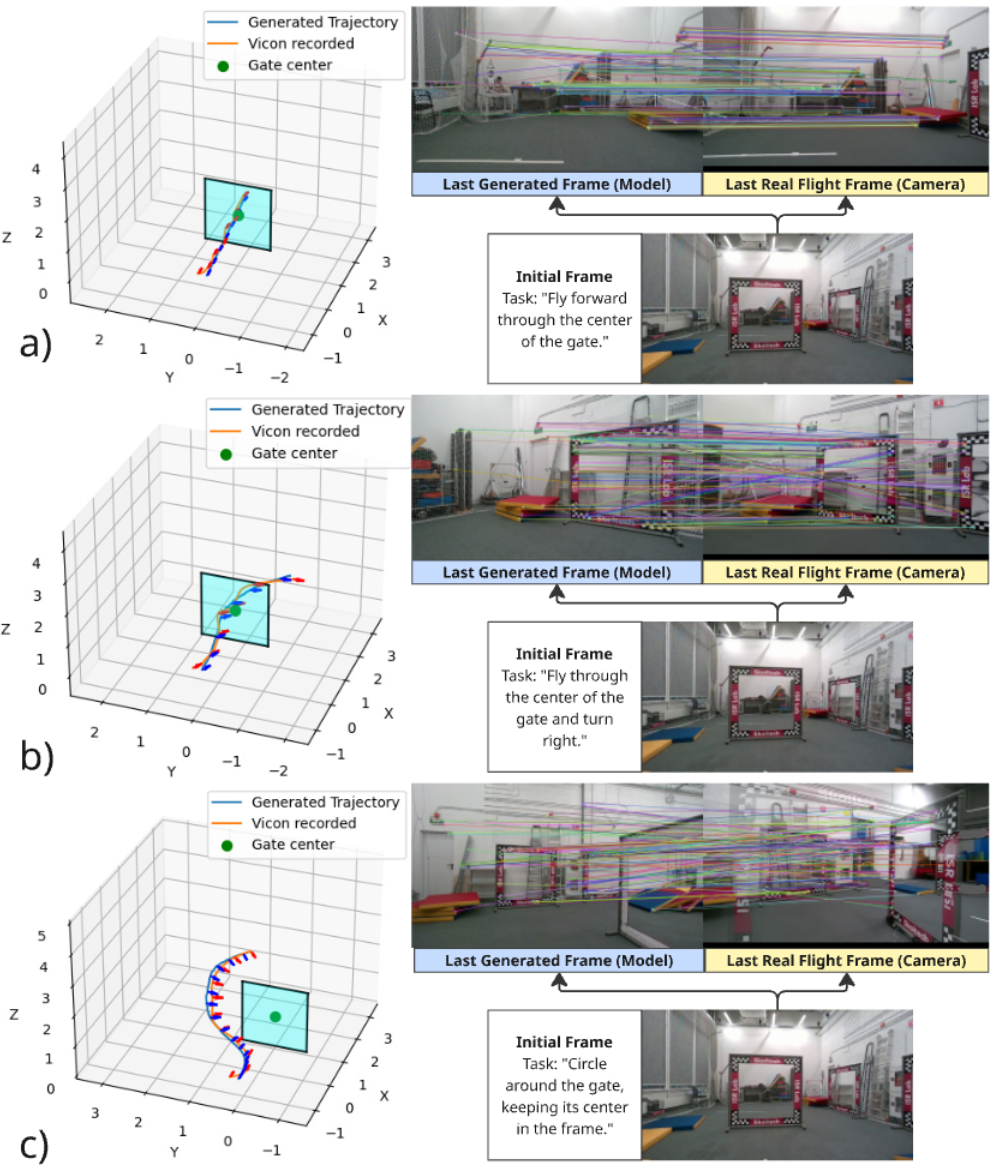}
\caption{Comparative analysis of drone flight trajectories. Left: 3D plots with orientation vectors of Vicon ground truth (blue) and trajectories estimated from synthetic video (red). Right: Final synthetic frame and the last frame from the corresponding real flight, annotated with ORB feature matching points. Flight maneuvers: (a) Direct flight through the gate center. (b) Center flight with a subsequent right turn. (c) Circumnavigation flight while maintaining gate observation.}
\label{fig:3_plots}
\end{figure}

A qualitative comparison showed strong visual consistency between the final frame of the generated video (Fig. \ref{fig:3_plots}), $I_{\text{gen}}[N]$, and the corresponding image captured by the onboard camera at task completion, $I_{\text{real}}[N]$. Both frames contained similar semantic cues and geometric structures. The differences that did appear were minor, mainly small orientation offsets ($\Delta \theta$) and the occasional presence of hallucinated objects in $I_{\text{gen}}[N]$ that were not present in the real environment an expected limitation of diffusion-based generative models. Overall, the close alignment suggests that the model was able to capture the underlying structure and dynamics of the flight task.

For the quantitative evaluation, we compared the trajectory generated by the pipeline with the ground-truth path measured by the VICON tracking system. Using a sequence of $K$ corresponding poses, the translational error at each step $k$ was defined as the Euclidean distance between the estimated position $\mathbf{p}_{\text{gen}}[k] = (x_k, y_k, z_k)$, while the rotational component was represented by its Tait-Bryan angles (yaw, pitch, roll) $\bm{\theta}_{\text{gen}}[k] = (\psi_k, \theta_k, \phi_k)$ and the ground-truth position $\mathbf{p}_{\text{gt}}[k] = (x_{gt,k}, y_{gt,k}, z_{gt,k})$ and orientation $\bm{\theta}_{\text{gen}}[k] = (\psi_{gt,k}, \theta_{gt,k}, \phi_{gt,k})$:
\begin{equation}
e_{\text{trans}}[k] = \left\| \mathbf{p}_{\text{gen}}[k] - \mathbf{p}_{\text{gt}}[k] \right\|_2
\end{equation}

The orientation error was quantified by calculating:

\begin{equation}
e_{\text{rot}}[k] = \left\| \bm{\theta}_{\text{gen}}[k] - \bm{\theta}_{\text{gt}}[k] \right\|_2
\end{equation}

\begin{table}[h!]
\centering
\caption{Quantitative results of trajectory accuracy. Values represent averages across three experimental trials.}
\label{tab:results}
\begin{tabular}{lccc}
\toprule
\textbf{Metric} & \textbf{Translation (\si{\meter})} & \textbf{Rotation (\si{\radian})} \\
\midrule
Mean Error & 0.25 & 0.19 \\
Max Error & 0.44 & 0.51 \\
RMSE & 0.28 & 0.24 \\
\bottomrule
\end{tabular}
\end{table}

The aggregated results from three experimental trials are presented in Table~\ref{tab:results}. The low values for all error metrics demonstrate that the generated trajectory closely approximated the actual flight path, with an average positional RMSE of \SI{0.28}{\meter} The orientation RMSE of \SI{0.23}{\radian}.

The small remaining errors can be traced to the sim-to-real gap and the stochastic nature of the diffusion process. Even so, achieving sub-meter accuracy demonstrates that the FlightDiffusion framework is capable of producing physically plausible and executable flight paths directly from high-level task descriptions.

\subsection{Simulation vs. Real-World Maneuver Evaluation}

\begin{figure}[t]
\centering
\includegraphics[width=1.0\linewidth]{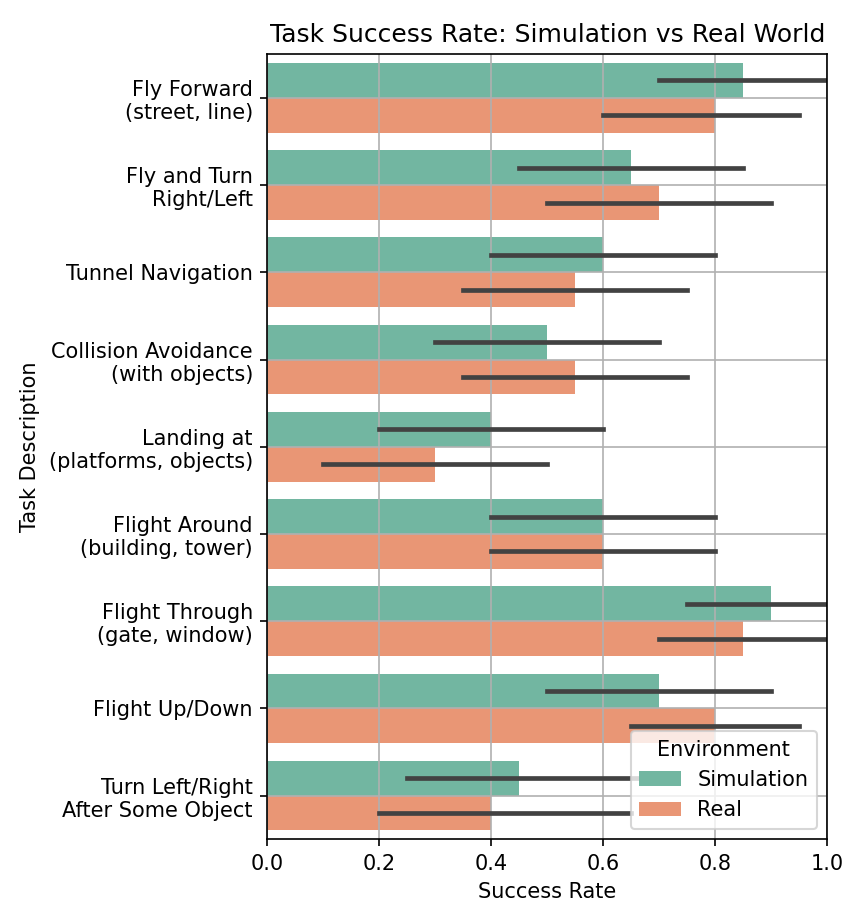}
\caption{Barplot comparing the success rates of flight maneuvers performed in simulation versus real-world conditions.}
\label{fig:real_vs_sim}
\end{figure}

\begin{figure*}[t!]
\centerline{\includegraphics[width=1.0\textwidth]{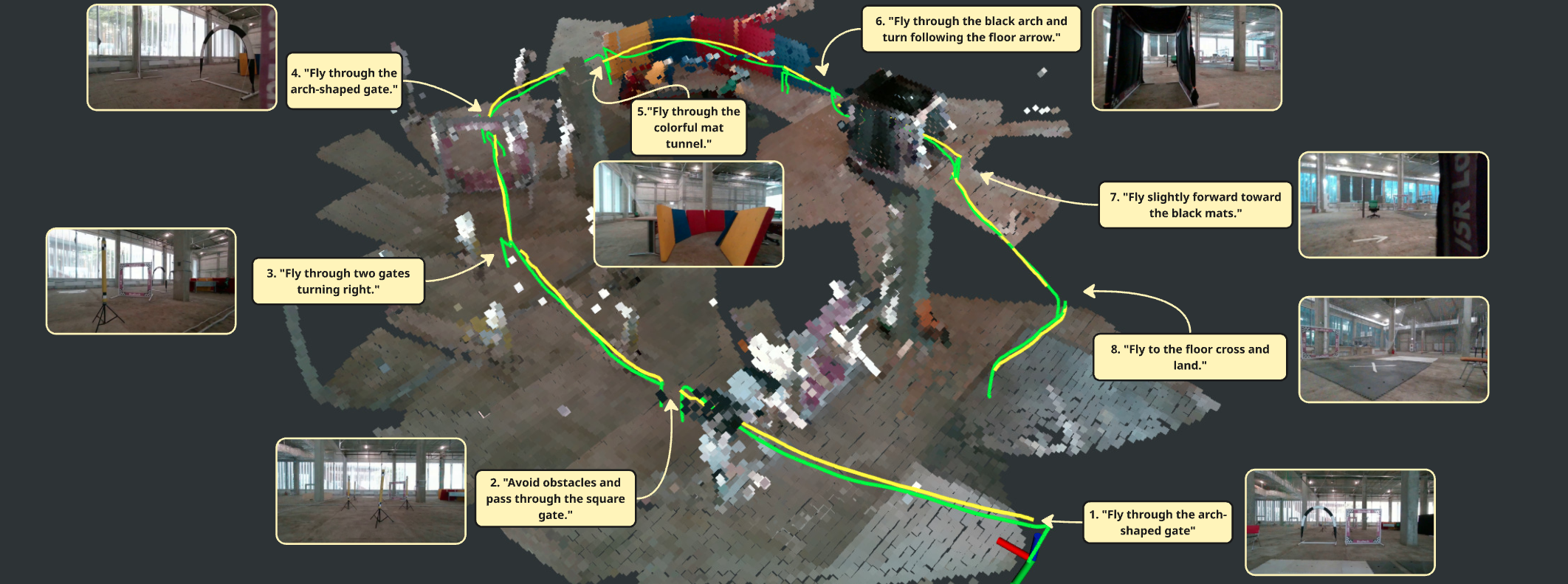}}
\caption{A reconstructed point cloud map of the flight environment, overlaid with trajectories from a complete long-horizon mission. The green line depicts the real-time trajectory recorded by the Visual Inertial Odometry system. The yellow lines represent the trajectories generated by the diffusion model at each key stage. The figure illustrates the iterative process where, for each subtask, the model uses a visual frame (inset) and a text description to generate a target path, which the drone then executes.}
\label{fig:long_term_exp_real}
\end{figure*}


To evaluate the transfer of autonomous flight performance from simulation to reality, each of nine distinct maneuvers (including forward flight, tunnel navigation, collision avoidance, and landing) were executed twenty times in both a simulated and a real-world environment, yielding a success rate for each condition (Fig. \ref{fig:real_vs_sim}). We used the small city world \cite{gazebo_worlds_oevreal} in the Gazebo simulation environment. A two-way Analysis of Variance (ANOVA) without replication was conducted to assess the effects of the environment (simulation vs. reality) and the maneuver type on the success rates. The analysis revealed no statistically significant main effect of the environment $(F(1, 16) = 0.394, p = 0.541)$, indicating that the overall mean success rate in simulation $(M = 0.628, SD = 0.162)$ was not significantly different from that in reality $(M = 0.617, SD = 0.177)$. However, a highly significant main effect of maneuver type was found $(F(8, 16) = 26.250, p < 0.001)$, confirming that the intrinsic difficulty varied substantially across the different flight tasks. A paired-sample t-test corroborated the ANOVA finding, showing no significant difference between the paired conditions $(t(8) = 0.318, p = 0.758)$. These results demonstrate that while the complexity of maneuvers significantly influences performance, the simulation serves as a highly valid environment, with overall system performance being statistically equivalent to real-world operation.

\subsection{Long Horizon Task}

\begin{figure}[t]
\centering
\includegraphics[width=0.8\linewidth]{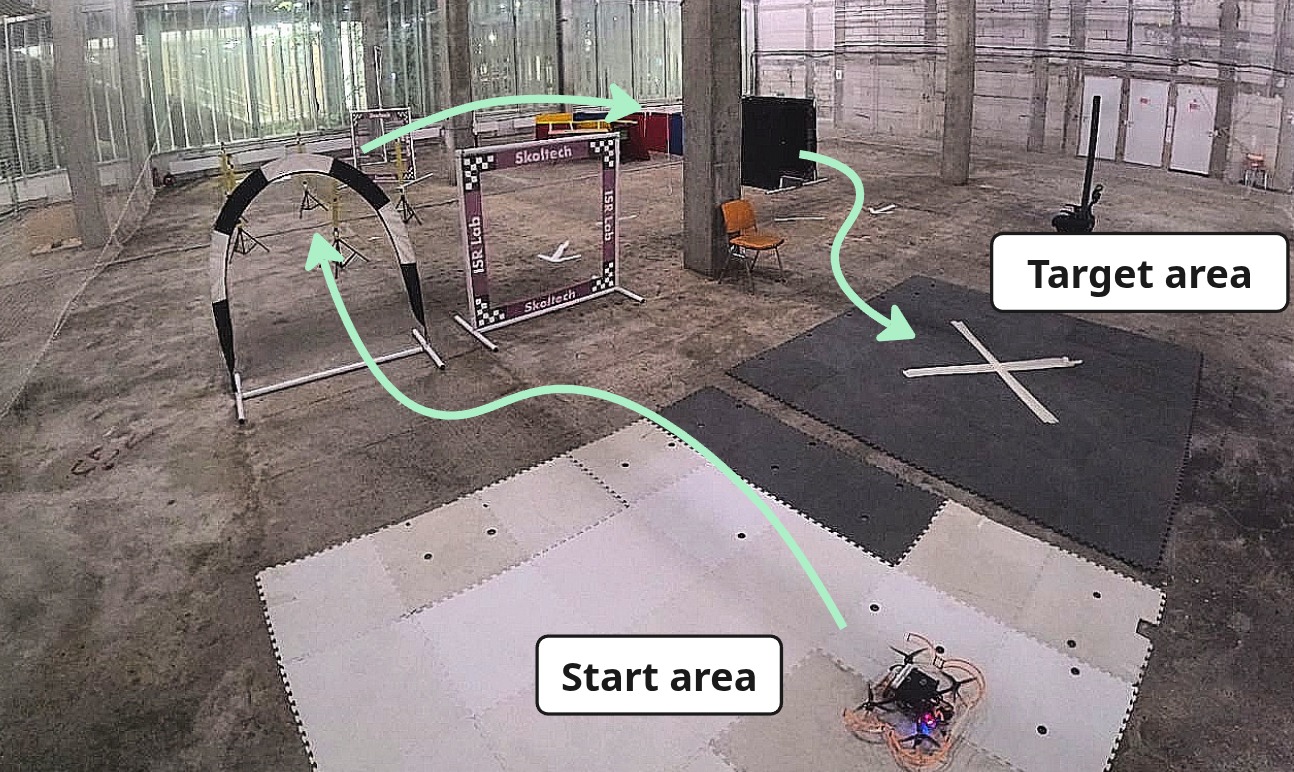}
\caption{The experimental drone platform in a controlled flight arena.}
\label{fig:real_setup_long}
\end{figure}

\begin{figure}[t]
\centering
\includegraphics[width=1.0\linewidth]{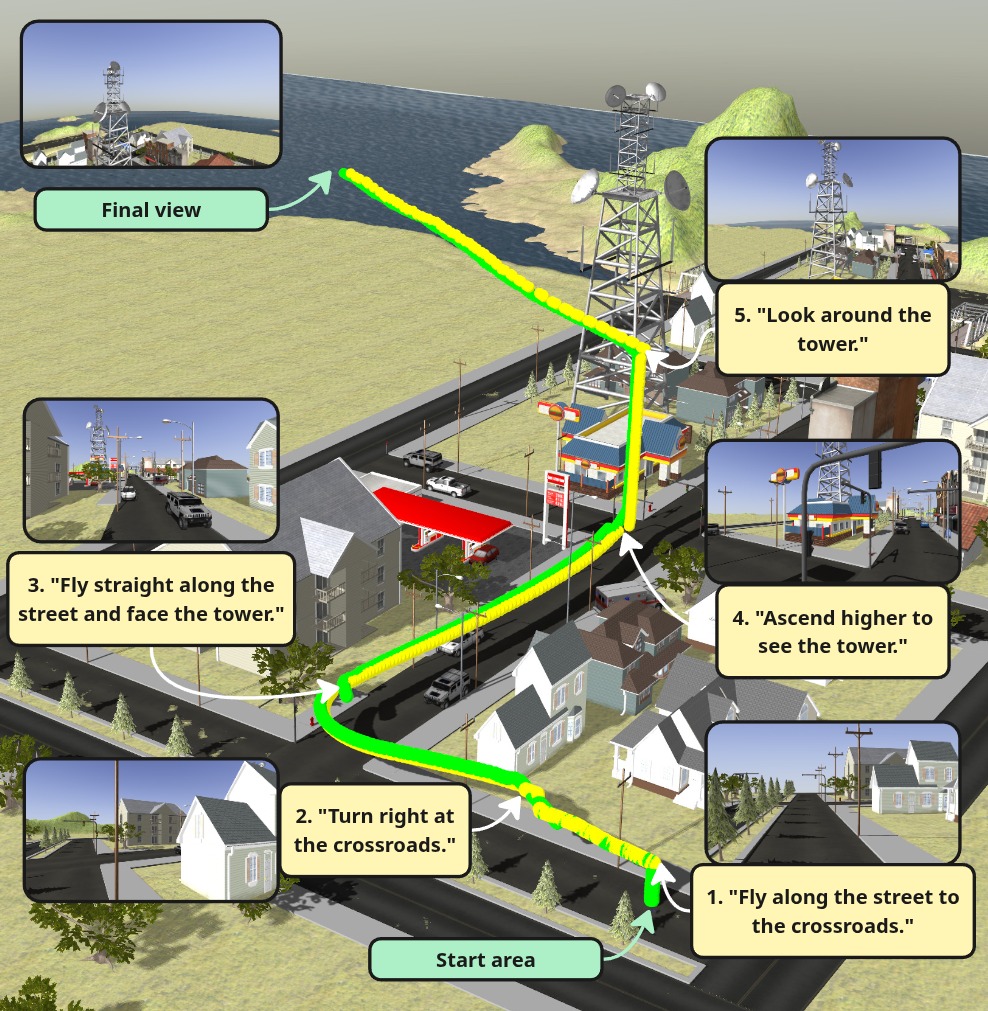}
\caption{Simulation environment for long-horizon tower inspection tasks. The green line represents the trajectory executed by ArduPilot SITL, while the yellow paths illustrate the generated trajectories. Input images and task specifications are shown for each planning stage.}
\label{fig:sim_exp}
\end{figure}

\textbf{Objective.} The primary objective of the experiment was to evaluate the drone's ability to autonomously complete a complex, long-horizon navigation course in simulated (Fig.\ref{fig:sim_exp}) and physical real-world (Fig.  \ref{fig:real_setup_long}) testing environments. 

\textbf{Global Task Definition.} The global long horizon task for  was defined as follows: 

\begin{itemize}
\item Simulated environment. "The drone navigates over the city, proceeds to the first crossroads, makes a right turn, locates the tower, ascends, and performs an inspection of the structure."
\item Real-world environment. "Starting from an initial position, navigate through two entrance gates, proceed through an arch, follow a forward path while avoiding collisions with yellow vertical obstacles, execute a smooth right turn to cross a square and a subsequent arch gate, fly between a set of colored mats, enter and traverse a black tunnel, follow the direction of an arrow marked on the ground, and finally land precisely on a white cross." 
\end{itemize}

\textbf{Setup.} The global task was provided to a high-level reasoning model. This model analyzed the long-horizon objective and, using the initial visual feed from the drone's camera, dynamically subdivided it into a sequence of short-horizon manageable subtasks. For instance, the initial task was decomposed into: Navigate to Arch Gate 1, Avoid Yellow Obstacles, Turn Right, and Traverse Square, etc. For each identified subtask (e.g.,``Fly through the arch-shaped gate"), the model performed a generalization and reasoning step based on the current image from the FPV camera (see Fig.\ref{fig:long_term_exp_real}). Given this image and the textual subtask prompt, the diffusion model generated a synthetic FPV video trajectory depicting the optimal flight path to complete the subtask. This generated video served as a visual prediction of the desired outcome, encoding not just a path but also the expected visual dynamics and drone orientation. The keyframes from the generated video trajectory were then processed. Upon completing a subtask (e.g., passing through the first arch gate), the drone's state was updated and a new image was taken. The drone switched to HOVER or LAND mode if the video generation was too long or the battery level became low. The reasoning model then acquired the next visual feed, identified the subsequent subtask from the global plan, and the process repeated: Reasoning → Video Generation → Path Planning → Execution. This iterative, closed-loop approach allowed the system to handle the entire long-horizon task by sequentially solving shorter-horizon problems, making it robust to minor positional drift and environmental changes.

\textbf{Results.} The FlightDiffusion framework proved effective, completing the full long-horizon mission for the real-world environment in 5 min and 20 sec at an average speed of 1 m/s, and 4 min for simulated environment, while each task required about 30 sec to generate the corresponding synthetic video. The diffusion model proved highly capable of generating realistic and physically plausible video trajectories that were successfully converted into flyable paths. (Fig.~\ref{fig:long_term_exp_real} and Fig.~\ref{fig:sim_exp}) illustrate the generated trajectory (yellow) and the trajectory performed by the UAV (green), shown in sequential order. A critical observation was the effectiveness of the hierarchical decomposition. By breaking down the problem, the model avoided the pitfalls of end-to-end planning over a long horizon, which often suffers from compounding errors. The generated trajectories for subtasks like the``smooth right turn" and ``flight between colored mats" were particularly notable for their natural and efficient kinematics.

\section{Conclusion and Future Work}

This paper introduced FlightDiffusion, a novel diffusion-model–based framework for autonomous drone navigation that unifies generative video synthesis, action-space generation, and policy learning within a single, cohesive architecture. Our results demonstrate that this paradigm represents a significant advancement in data-driven aerial robotics, effectively addressing the critical challenges of data scarcity and the sim-to-real transfer gap.

The core strength of FlightDiffusion lies in its dual capability. Primarily, it functions as a powerful policy learning tool, enabling reasoning-driven navigation from a single FPV frame. Secondly, and perhaps more impactfully, it serves as a highly effective data generation engine. The framework's ability to synthesize vast quantities of physically plausible and diverse FPV trajectories, complete with accurate state–action pairs, provides a scalable and cost-effective solution for creating large-scale training datasets. This capability was quantitatively validated by the low error metrics between generated and actual flight paths, with a positional RMSE of 0.28 m and a orientation RMSE of 0.24 rad, confirming the executability of the synthesized data.

Crucially, our extensive evaluation establishes a robust case for the validity of this approach. The statistical equivalence in performance between simulated and real-world environments, as confirmed by a non-significant ANOVA result $(F(1, 16) = 0.394, p = 0.541)$ and a paired t-test $(t(8) = 0.318, p = 0.758)$, demonstrates an exceptional level of sim-to-real transfer. This indicates that policies trained on FlightDiffusion's synthetic data possess the robustness and generalizability required for deployment in the real world. The significant effect of maneuver type $(F(8, 16) = 26.250, p < 0.001)$ further confirms that the framework accurately captures the intrinsic difficulty of various navigation tasks, rather than simply masking failure modes.

Furthermore, the framework's efficacy in long-horizon mission completion, achieved through a hierarchical decomposition strategy, highlights its practical utility for complex, real-world applications. By mitigating compounding errors through subtask generation, the model produced kinematically natural and efficient trajectories.

In summary, FlightDiffusion successfully establishes diffusion-based reasoning as a promising and powerful paradigm for aerial robotics. It moves beyond mere imitation learning by providing a generative foundation for both perception and action. The released datasets and the validated methodology provide a valuable resource for future research, paving the way for more scalable, robust, and intelligent autonomous systems capable of learning from and reasoning about dynamic environments with unprecedented efficiency. Future work will focus on expanding the diversity of environmental conditions and integrating more complex multi-modal inputs to further enhance generalizability.

\bibliographystyle{IEEEtran}
\bibliography{ref}

\end{document}